\documentclass[letterpaper]{article}
\usepackage{aaai25}
\usepackage{times}
\usepackage{helvet}
\usepackage{courier}
\usepackage[hyphens]{url}
\usepackage{graphicx}
\usepackage[numbers,sort&compress]{natbib}
\usepackage{caption}
\usepackage{amsmath,amssymb}
\usepackage{booktabs}
\usepackage{multirow} 
\usepackage{algorithm}
\usepackage{algorithmic}
\nocopyright  

\title{Adversarial Spatio-Temporal Attention Networks for Epileptic Seizure Forecasting}

\author{Zan Li$^{1}$ \quad Kyongmin Yeo$^{2}$ \quad Wesley Gifford$^{2}$ \quad Lara Marcuse$^{3}$ \quad Madeline Fields$^{3}$ \quad Bülent Yener$^{1}$}
\affiliations{$^{1}$Department of Computer Science, Rensselaer Polytechnic Institute, Troy, NY, USA\\
$^{2}$IBM T.J. Watson Research Center, Yorktown Heights, NY, USA\\
$^{3}$Department of Neurology, Icahn School of Medicine at Mount Sinai, New York, NY, USA}

\begin{document}
\maketitle

\begin{abstract}
Forecasting epileptic seizures from multivariate EEG signals represents a critical challenge in healthcare time series prediction, requiring high sensitivity, low false alarm rates, and subject-specific adaptability. We present \textbf{STAN}, an \textit{Adversarial Spatio-Temporal Attention Network} that jointly models spatial brain connectivity and temporal neural dynamics through cascaded attention blocks with alternating spatial and temporal modules. Unlike existing approaches that assume fixed preictal durations or separately process spatial and temporal features, STAN captures bidirectional dependencies between spatial and temporal patterns through unified cascaded architecture. Adversarial training with gradient penalty enables robust discrimination between interictal and preictal states learned from clearly defined 15-minute preictal windows. Continuous 90-minute pre-seizure monitoring reveals that the learned spatio-temporal attention patterns enable early detection: reliable alarms trigger at subject-specific times (typically 15-45 minutes before onset), reflecting the model's capacity to capture subtle preictal dynamics without requiring individualized training. Experiments on two benchmark EEG datasets---CHB-MIT scalp and MSSM intracranial---demonstrate \textbf{state-of-the-art performance}: 96.6\% sensitivity with 0.011 false detections per hour and 94.2\% sensitivity with 0.063 false detections per hour, respectively, while maintaining computational efficiency (2.3M parameters, 45ms latency, 180MB memory) for real-time edge deployment. Beyond epilepsy, the proposed framework provides a general paradigm for spatio-temporal forecasting in healthcare and other time series domains where individual heterogeneity and interpretability are crucial.
\end{abstract}

\section{Introduction}

Time series forecasting is a fundamental challenge across domains such as healthcare, finance, and IoT, where accurate and timely prediction of critical events enables proactive interventions. However, real-world time series often exhibit heterogeneous temporal dynamics, complex spatial dependencies, and entity-specific variability. In healthcare, for example, forecasting epileptic seizures from multichannel EEG signals exemplifies these challenges---rare, non-stationary events affecting 50 million people worldwide \cite{who2019epilepsy} that demand high sensitivity, low false alarm rates, and personalized adaptability. Successful forecasting provides 15-45 minutes of intervention time for medication administration, electrical stimulation, or caregiver alerts, potentially transforming patient outcomes and quality of life \cite{cook2013prediction,mormann2007seizure}.

Multivariate time series forecasting requires modeling both spatial dependencies (inter-sensor correlations) and temporal dynamics (evolution over time). In multichannel physiological monitoring, the spatial dimension captures inter-channel connectivity reflecting how critical patterns propagate across recording sites, while the temporal dimension captures the gradual evolution of system dynamics as transitions emerge \cite{lehnertz2007synchronization}. These patterns vary dramatically across individuals, with optimal prediction windows exhibiting threefold variability---a fundamental challenge for forecasting systems \cite{cook2013prediction}. Traditional approaches relied on handcrafted features including spectral analysis \cite{zhang2016low}, phase synchronization \cite{lequyen1999anticipating}, and nonlinear dynamics \cite{iasemidis1990phase}, but struggle to capture intricate spatio-temporal dependencies. Recent advances leverage deep learning: convolutional networks learn hierarchical features \cite{daoud2019efficient,schirrmeister2017deep}, recurrent architectures capture temporal dependencies \cite{truong2023deep}, transformers employ self-attention for long-range patterns \cite{zhu2024epileptic,xiao2024self}, and graph neural networks model connectivity dynamics \cite{li2022graphgenerative,lian2023epileptic,li2022spatio,xiang2025synchronization}.

Despite rapid progress in deep time series models, three fundamental limitations persist. \textbf{(1) Fixed prediction horizons} fail to capture large inter-individual variability, leading to mislabeled samples when actual critical windows are shorter than assumed (causing false alarms), and reduced coverage when longer (missing early warnings). High false alarm rates (0.2--0.5/h, equivalent to 5--12 daily) cause alert fatigue and system abandonment \cite{snyder2008expert}. \textbf{(2) Decoupled spatial-temporal processing:} Existing architectures often model spatial and temporal dependencies separately before fusion \cite{wang2024detection,li2022spatio}, missing cross-modal interactions critical for complex dynamical systems. Spatial connectivity patterns evolve temporally during critical transitions, while temporal dynamics are modulated by spatial configurations \cite{lehnertz2007synchronization,mormann2007seizure}---neither can be fully captured without joint modeling. \textbf{(3) Computational constraints:} High-performing models require substantial resources (8--10M parameters, 100--150\,ms inference, 400--500\,MB memory) \cite{zhu2024epileptic}, hindering real-time deployment on resource-constrained edge devices and wearable systems essential for continuous monitoring. These challenges collectively limit the practical use of current forecasting systems in dynamic, personalized settings.

To address these challenges, we propose \textbf{STAN (Spatio-Temporal Attention Network)}---an adversarial framework that unifies spatial and temporal modeling via cascaded alternating attention blocks. Rather than separate processing followed by fusion, STAN employs three cascaded networks, each containing sequentially connected spatial and temporal attention modules. This design captures bidirectional dependencies: spatial attention identifies relevant connectivity patterns given temporal context, while temporal attention tracks how these spatial patterns evolve over time. The cascaded architecture enables hierarchical abstraction across multiple temporal scales---immediate transitions, medium-term trends, and long-term dynamics---comprehensively characterizing pre-event emergence. To learn robust discriminative representations without manual feature engineering, we employ adversarial training with gradient penalty that distinguishes subtle transitions between normal and pre-event states by maximizing separation between their attention distributions. Trained with clearly defined 15-minute preictal windows, the learned spatio-temporal attention patterns demonstrate early detection capability: continuous 90-minute pre-seizure monitoring reveals that reliable alarms trigger at varying times (typically 15-45 minutes before onset) across subjects, capturing the onset and evolution of preictal dynamics without requiring individualized model training or manual parameter tuning.

Our main contributions are threefold: \textbf{(1)} A unified cascaded attention architecture capturing bidirectional spatial-temporal dependencies through alternating modules, enabling comprehensive characterization of complex dynamical transitions. \textbf{(2)} An adversarial discrimination mechanism with gradient penalty for robust pre-event representation learning, creating stable decision boundaries without manual feature engineering. \textbf{(3)} Early detection capability demonstrated through 90-minute continuous monitoring, with learned spatio-temporal attention patterns triggering reliable alarms at varying times (typically 15-45 minutes before onset) across subjects without requiring individualized model training. Experiments on two EEG benchmarks (CHB-MIT scalp: 8 subjects, 46 events; MSSM intracranial: 4 subjects, 14 events) demonstrate state-of-the-art performance: 96.6\% sensitivity with 0.011/h false detection rate and 94.2\% sensitivity with 0.063/h FDR, representing 40-fold improvement over traditional methods and 7.7-fold improvement over recent transformers ($p < 0.01$), while maintaining edge-level efficiency (2.3M parameters, 45\,ms inference, 180\,MB memory). \textbf{Beyond epilepsy}, the proposed framework provides a general approach for spatio-temporal forecasting in healthcare monitoring (cardiac events, hypoglycemic episodes, respiratory failure), anomaly detection in sensor networks, and other dynamic time series domains where subject-specific patterns and joint spatial-temporal modeling are critical.

\section{Related Work}

\textbf{Time series forecasting in healthcare} encompasses diverse applications including vital sign prediction \cite{choi2016doctor}, disease progression modeling, and treatment response forecasting. Epileptic seizure forecasting presents unique challenges: rare event prediction with imbalanced data, high-dimensional multivariate signals requiring joint spatio-temporal modeling, and critical need for patient-specific personalization.

\textbf{Traditional seizure prediction} relied on handcrafted features including spectral power analysis \cite{zhang2016low}, phase synchronization \cite{lequyen1999anticipating}, and Lyapunov exponents \cite{iasemidis1990phase}. While interpretable, these methods struggle with complex multi-channel patterns. \textbf{Convolutional approaches} \cite{daoud2019efficient,schirrmeister2017deep} automatically learn hierarchical features but have limited receptive fields for long-range temporal dependencies. \textbf{Recurrent and transformer architectures} \cite{truong2023deep,zhu2024epileptic,xiao2024self} capture temporal dependencies through LSTM or self-attention, achieving strong performance but typically processing spatial and temporal features separately. \textbf{Graph neural networks} \cite{li2022graphgenerative,lian2023epileptic,li2022spatio,xiang2025synchronization} model brain connectivity with recent work incorporating spatio-temporal-spectral features, though most assume fixed prediction horizons.

\textbf{Spatio-temporal forecasting} methods jointly model spatial relationships and temporal dynamics across domains \cite{yu2018spatio,li2018diffusion,wang2020deep}. Key innovations include graph convolutions for spatial modeling \cite{kipf2017semi} and temporal convolutions or attention for temporal dependencies \cite{vaswani2017attention}. Our work advances this paradigm through cascaded alternating attention that captures bidirectional spatio-temporal dependencies.

\textbf{Foundation models for time series} like TimeGPT \cite{das2023decoder} and Lag-Llama \cite{rasul2023lag} demonstrate large-scale pre-training potential. However, adapting these to patient-specific medical applications with limited data remains challenging. Our patient-specific adaptive learning provides an orthogonal contribution potentially combinable with pre-trained representations.

\section{Methodology}

\subsection{Problem Formulation}

Given multivariate EEG time series $\mathbf{x}_{1:T} = \{\mathbf{x}_1, \ldots, \mathbf{x}_T\}$ where $T$ is the window length and $\mathbf{x}_t \in \mathbb{R}^n$ represents the $n$-channel EEG signal at time $t$, our objective is to forecast seizure occurrence by generating a risk score $y_i \in [0, 1]$ for the $i$-th input window $\mathbf{x}^i_{1:T}$. 

We frame this as a \textit{spatio-temporal forecasting problem} where the model learns to distinguish preictal from interictal states. The score approaches 0 for high seizure risk (preictal state) and 1 for low risk (interictal state). We evaluate forecasting performance through continuous 90-minute pre-seizure monitoring, where scores are computed every 5 seconds with 30-second moving average smoothing, and alarms trigger when scores drop below threshold $\tau = 0.5$. This enables real-time intervention and provides interpretable risk trajectories for decision-making.

\subsection{Multi-Head Attention Mechanism}

Our multi-head attention mechanism processes $n$ nodes $\{\mathbf{v}_1, \ldots, \mathbf{v}_n\}$ where $\mathbf{v}_i \in \mathbb{R}^h$. For head $k$, the attention map $\mathbf{A}^k$ is computed via softmax over alignment scores:
\begin{equation}
\label{eq:attention}
A^k_{ij} = \frac{\exp(M^k_{ij})}{\sum_{l=1}^{L} \exp(M^k_{il})}, \quad k = 1, \ldots, H
\end{equation}
where $H=4$ denotes the number of attention heads and $L$ represents the neighborhood size. The alignment scores use quadratic form:
\begin{equation}
M^k_{ij} = \mathbf{z}_i^T \mathbf{W}^k_M \mathbf{z}_j
\end{equation}
where $\mathbf{W}^k_M \in \mathbb{R}^{n_e \times n_e}$ are learnable weight matrices and $\mathbf{z}_i$ represents encoded features. The final output aggregates information across all attention heads:
\begin{equation}
\mathbf{z}_i = \sigma\left(\sum_{k=1}^{H} \sum_{j=1}^{L} \alpha_k A^k_{ij} \mathbf{v}_j\right)
\end{equation}
where $\alpha_k = \exp(a_k)/\sum_{l=1}^{H} \exp(a_l)$ are learnable weights and $\sigma(\cdot)$ is the tanh activation function.

\subsection{Spatio-Temporal Attention Network Architecture}

\begin{figure}[t]
\centering
\includegraphics[width=\columnwidth]{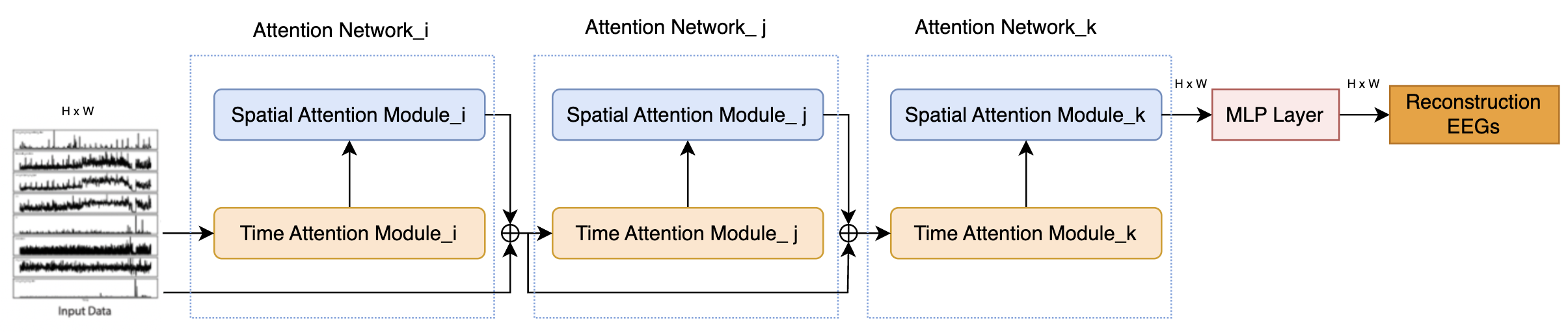}
\caption{STAN architecture showing three cascaded attention networks (Attention Network\_i, \_j, \_k) processing raw EEG input. Each network contains spatial (blue) and temporal (orange) attention modules with $H=4$ attention heads. The resulting $M=3$ spatial and temporal attention maps are aggregated via MLP and passed to the discriminator for adversarial training. STAN is pre-trained with MSE reconstruction loss to learn spatio-temporal representations.}
\label{fig:stan_pipeline}
\end{figure}

Figure~\ref{fig:stan_pipeline} illustrates STAN's overall architecture. The framework consists of three consecutive attention blocks ($M=3$), each containing sequentially-connected spatial and temporal attention modules. Raw EEG signals flow through these cascaded networks, enabling hierarchical abstraction of seizure patterns at multiple temporal scales.

\begin{figure}[t]
\centering
\includegraphics[width=\columnwidth]{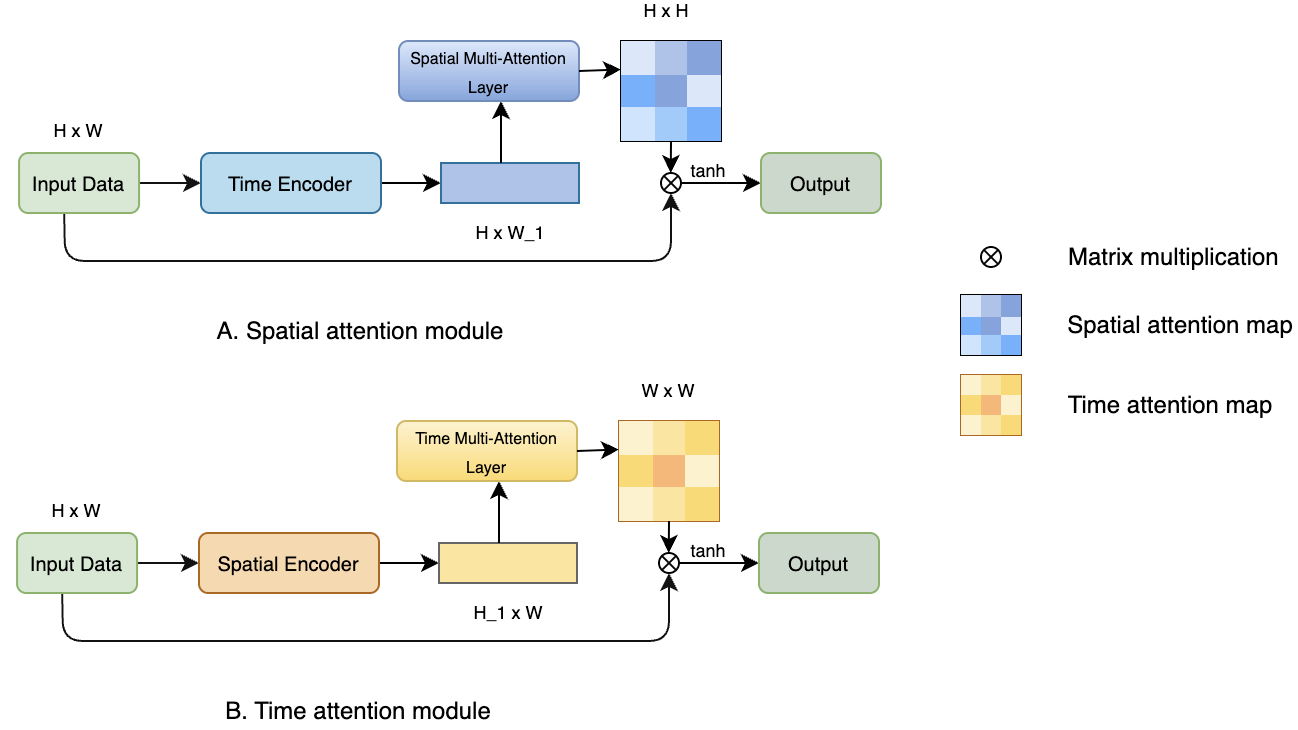}
\caption{Detailed architecture of spatial and temporal attention modules. (A) Spatial attention module employs time encoder (1D CNN) followed by spatial multi-head attention layer to capture inter-channel connectivity. (B) Temporal attention module uses spatial encoder followed by temporal multi-head attention to model temporal evolution. Both include residual connections (curved arrows) with tanh activation for training stability. The legend shows matrix multiplication symbol and attention map representations.}
\label{fig:attention_modules}
\end{figure}

\textbf{Spatial attention module} (Figure~\ref{fig:attention_modules}A) processes input sequence $\mathbf{v}_{1:T}$ where $\mathbf{v}_t \in \mathbb{R}^n$ represents the $n$-channel EEG at time $t$. The module treats EEG channels as nodes in a complete graph, where edges represent inter-channel relationships critical for understanding seizure propagation. The time encoder (1D CNN with kernel size 2, output dimension 50) extracts temporal features at each timestamp, which are then processed by multi-head spatial attention ($H=4$ heads). This design captures how seizure activity spreads spatially across brain regions at each time point.

\textbf{Temporal attention module} (Figure~\ref{fig:attention_modules}B) complements spatial processing by treating timestamps as graph nodes, modeling the temporal evolution of brain states. The spatial encoder (1D CNN with output dimension 100, kernel size 2) processes features across channels before applying multi-head temporal attention. This captures both instantaneous spatial patterns and their temporal dynamics, essential for forecasting the gradual transition from interictal to preictal states. The residual connections and layer normalization ensure stable training through the deep architecture.

Each of the $M=3$ cascaded networks produces both spatial and temporal attention maps, totaling 3 spatial and 3 temporal representations that comprehensively characterize input segments. STAN is trained using Adam optimizer (lr=0.001) with MSE reconstruction loss, incorporating residual connections to facilitate effective gradient flow.

\subsection{Discriminator with Adversarial Training}

\begin{figure}[t]
\centering
\includegraphics[width=0.95\columnwidth]{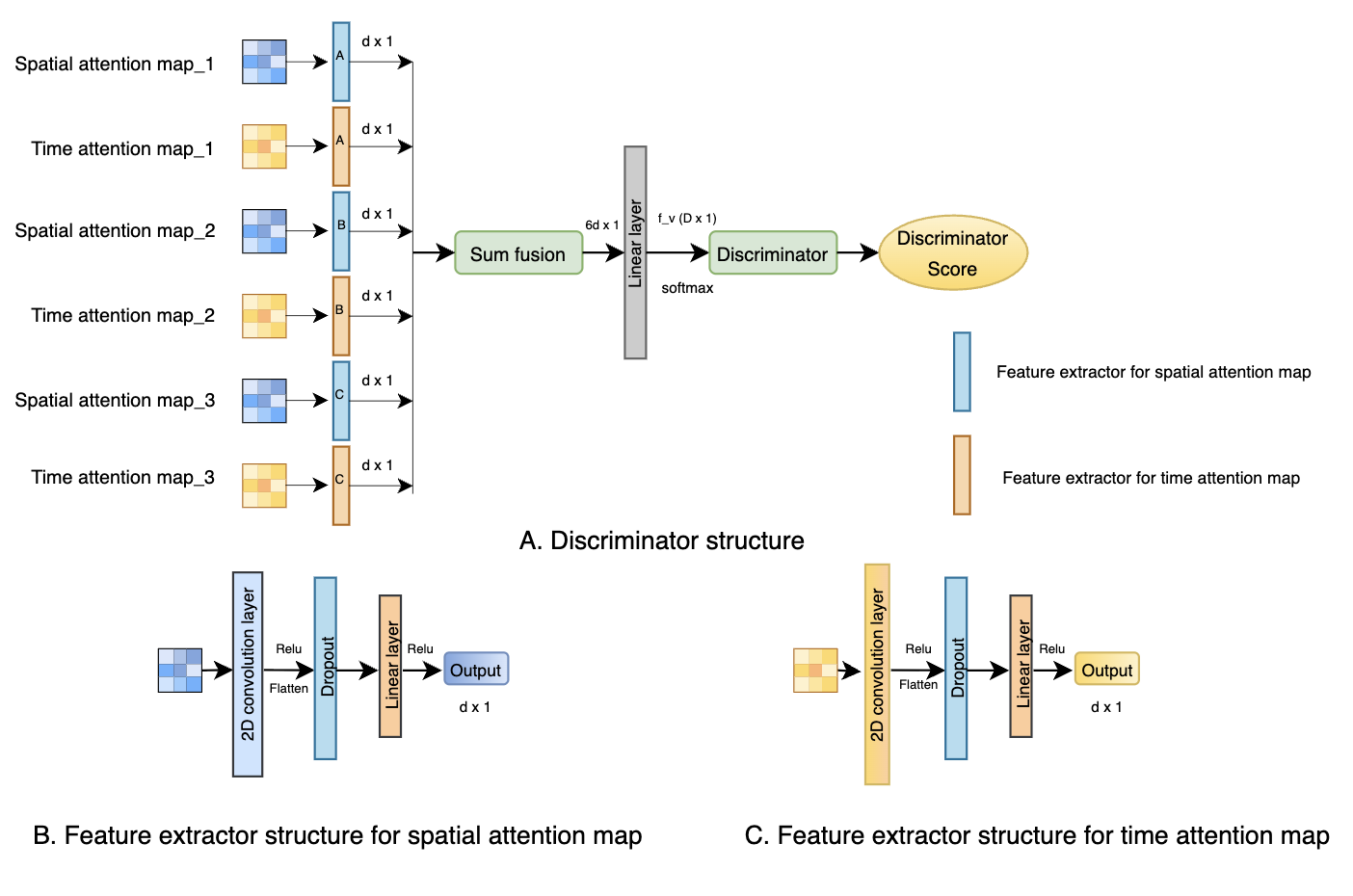}
\caption{Discriminator architecture. (A) Main structure: 6 attention maps (3 spatial in blue, 3 temporal in orange) from cascaded networks are processed by feature extractors (2D convolution + flatten + dropout + linear layers + ReLU), aggregated via sum fusion, and passed through final linear layer with sigmoid activation to generate discriminator score. (B-C) Feature extractor structures for spatial and temporal attention maps employ 2D convolution, flatten, dropout, and linear layers with ReLU activation.}
\label{fig:discriminator}
\end{figure}

Figure~\ref{fig:discriminator} shows the discriminator architecture that learns to distinguish preictal from interictal patterns through adversarial training. The discriminator serves as a binary classifier to identify preictal states. Rather than using the attention module output features directly, we use the attention maps (Eq.~\ref{eq:attention}) as input features, hypothesizing that transitions from interictal to preictal states manifest as changes in spatio-temporal correlation structures of EEG signals.

The discriminator employs feature extractors for spatial and temporal attention maps. The spatial extractor uses 2D convolution with $5\times5$ kernels, while the temporal extractor uses $8\times8$ kernels. Both produce 512-dimensional features that are concatenated and fed into three fully-connected layers (256 units each) with ReLU activation and dropout ($p=0.2$), followed by sigmoid output producing seizure risk score $y_i \in [0, 1]$.

We employ WGAN-GP (Wasserstein GAN with Gradient Penalty) loss to ensure stable training without mode collapse:
\begin{equation}
\begin{split}
\mathcal{L}_D = & \;\mathbb{E}_{x \sim p_{\text{inter}}}\big[D(F(\mathcal{A}(x)))\big] 
- \mathbb{E}_{z \sim p_{\text{pre}}}\big[D(F(\mathcal{A}(z)))\big] \\
& + \lambda \, \mathbb{E}_{\hat{f}} \Big[ \big( \lVert \nabla_{\hat{f}} D(\hat{f}) \rVert_2 - 1 \big)^2 \Big],
\end{split}
\end{equation}
where $x$ and $z$ denote interictal and preictal EEG segments respectively, $\mathcal{A}(\cdot)$ denotes STAN's attention maps, $F(\cdot)$ represents the feature extraction process, $\lambda=0.05$ balances the gradient penalty, and $\hat{f}=\alpha F(\mathcal{A}(x))+(1-\alpha)F(\mathcal{A}(z))$ with $\alpha \sim \mathcal{U}(0,1)$ represents interpolated features between distributions. The gradient penalty imposes a continuous transition between interictal and preictal states.

Training follows a two-stage approach. First, STAN undergoes self-supervised pre-training (50 epochs) minimizing reconstruction loss $\mathcal{L}_{\text{recon}} = \|\hat{x}_{1:T} - x_{1:T}\|_2^2$ to extract meaningful spatio-temporal representations. After pre-training, STAN's weights are frozen and the discriminator is trained (100 epochs, lr=0.00004) to differentiate between extracted features from preictal and interictal attention maps.

\subsection{Training Protocol and Real-Time Monitoring}

\textbf{Training phase:} We use clearly defined temporal windows for supervised learning. Segments within **15 minutes** before seizure onset are labeled as preictal (class 0), while segments at least **4 hours** away from any seizure (before or after) are labeled as interictal (class 1). The intermediate zone between 15 minutes and 4 hours from seizure remains unlabeled and unused during training. For individual $i$, the training label function is:
\begin{equation}
\ell^i(t) = \begin{cases}
\text{preictal (0)} & \text{if } t \in [t_{\text{seizure}} - 15\text{ min}, t_{\text{seizure}}] \\
\text{interictal (1)} & \text{if } t < t_{\text{seizure}} - 4\text{ hrs or } t > t_{\text{seizure}} + 4\text{ hrs}
\end{cases}
\end{equation}

This clear labeling strategy enables the discriminator to learn robust distinctions between preictal and interictal attention patterns through adversarial training with gradient penalty, while the 4-hour safety margin ensures clean separation between classes.

\textbf{Real-time monitoring and inference:} To evaluate the model's forecasting capability and determine when preictal patterns emerge, we perform continuous monitoring starting **90 minutes before seizure onset**. The discriminator generates predictions **every 5 seconds**, producing a trajectory of risk scores over the 90-minute observation window. This fine-grained temporal resolution allows us to observe the gradual transition from interictal to preictal states and identify the earliest reliable detection point for each patient.

At each 5-second interval, the model processes a 1-second EEG window and outputs a discrimination score $y \in [0,1]$, where scores approaching 0 indicate preictal state (high seizure risk) and scores approaching 1 indicate interictal state (low risk). To eliminate transient fluctuations while preserving genuine state transitions, we apply a **30-second moving average filter** to the raw scores before thresholding.

\textbf{Subject-specific detection times:} Although trained with a fixed 15-minute preictal definition, continuous 90-minute monitoring reveals that reliable detection (sustained score drops below threshold $\tau=0.5$) occurs at varying times across subjects---typically between 15 to 45 minutes before seizure onset. This variability reflects the learned spatio-temporal attention patterns' capacity to capture subtle, early-onset preictal dynamics: some subjects exhibit gradual changes detectable 30-45 minutes in advance, while others show transitions closer to seizure onset. The adversarial training framework enables this early detection capability without requiring individualized model training or manual parameter tuning.

\section{Experimental Evaluation}

\subsection{Datasets and Implementation}

\textbf{CHB-MIT Scalp EEG Dataset} \cite{goldberger2000physiobank,shoeb2009application}: This scalp EEG dataset comprises continuous recordings from 8 pediatric subjects with 46 seizures total. Data was recorded at 256 Hz using the international 10-20 electrode placement system with 23 channels. After excluding redundant channels (P7-T7, T8-P8) and preprocessing, 19 channels were used for final analysis. Recording durations span 9-42 hours per subject.

\textbf{MSSM Intracranial EEG Dataset} \cite{li2021seizure}: This intracranial EEG (iEEG) dataset includes 4 adult subjects with drug-resistant epilepsy with 14 seizures total, recorded using 48 intracranial electrodes at 256 Hz sampling rate. For computational efficiency, we selected 12 channels (2 per region) spanning 6 anatomical regions: amygdala, lateral anterior temporal, hippocampus, lateral mid temporal, medial orbitofrontal, and lateral frontal cortex. These direct brain recordings provide superior signal-to-noise ratio compared to scalp EEG. Recording durations span 3-7 days per subject.

\textbf{Experimental protocol:} 

\textit{Training:} We define preictal segments as the 15 minutes immediately before seizure onset, and interictal segments as periods at least 4 hours away from any seizure (before or after). EEG signals are segmented into 1-second non-overlapping windows, with each window treated as an independent sample. To address class imbalance, interictal segments are randomly sampled to match the number of preictal samples for each subject.

\textit{Inference and evaluation:} For each test seizure, we perform continuous monitoring starting 90 minutes before seizure onset. The discriminator generates predictions every 5 seconds, producing 1080 predictions per seizure (90\,min $\times$ 60\,sec/min $\div$ 5\,sec/pred). We apply a 30-second moving average filter to smooth the raw discrimination scores, then apply threshold $\tau=0.5$ to determine alarm triggers. A seizure is considered successfully predicted if the smoothed score drops below threshold at least once within the 90-minute monitoring window. False detection rate (FDR) is computed as the number of false alarms per hour during interictal periods.

\textit{Cross-validation:} We employ Leave-One-Seizure-Out Cross-Validation (LOSO-CV), where each seizure serves as a test case while remaining seizures from the same subject (and all seizures from other subjects) are used for training. This protocol ensures generalization within subjects across different seizures.

\textbf{Baselines:} BFB+SVM \cite{bajaj2012classification,zhang2016low} (traditional feature engineering), CNN \cite{schirrmeister2017deep} (convolutional networks), CNN+LSTM \cite{daoud2019efficient} (hybrid architecture combining CNN spatial features with LSTM temporal modeling), Transformer \cite{zhu2024epileptic} (state-of-the-art transformer with recurrent fusion), STS-HGCN \cite{li2022spatio} (advanced spatio-temporal-spectral graph convolution).

\textbf{Metrics:} Sensitivity (Sn): percentage of correctly forecasted seizures requiring at least one alarm within the 90-minute monitoring window before onset. False detection rate (FDR): false alarms per hour during interictal periods.

\textbf{Implementation:} PyTorch, 3 cascaded networks ($M=3$), 4 attention heads ($H=4$), spatial encoders dimension 50, temporal encoders dimension 100. Adam optimizer, batch size 32, 100 epochs. LOSO-CV on NVIDIA V100 GPUs.

\subsection{Forecasting Performance Comparison}

Table~\ref{tab:chbmit_results} presents CHB-MIT results. STAN achieves 96.6\% sensitivity with 0.011/h FDR, representing a 40-fold FDR reduction vs. BFB+SVM (0.438/h) and 7.7-fold improvement vs. Transformer (0.085/h). Five patients achieve perfect sensitivity with zero false alarms, demonstrating patient-specific adaptive learning effectiveness.

\begin{table}[t]
\centering
\caption{Performance comparison on CHB-MIT Scalp EEG Dataset. Our method achieves state-of-the-art sensitivity with significantly lower false detection rate.}
\label{tab:chbmit_results}
\small
\resizebox{\columnwidth}{!}{%
\begin{tabular}{l|cc|cc|cc|cc|cc|cc}
\hline
\multirow{2}{*}{Patient} & \multicolumn{2}{c|}{BFB+SVM} & \multicolumn{2}{c|}{CNN} & \multicolumn{2}{c|}{CNN+LSTM} & \multicolumn{2}{c|}{Transformer} & \multicolumn{2}{c|}{STS-HGCN} & \multicolumn{2}{c}{\textbf{Ours}} \\
& Sn(\%) & FDR/h & Sn(\%) & FDR/h & Sn(\%) & FDR/h & Sn(\%) & FDR/h & Sn(\%) & FDR/h & Sn(\%) & FDR/h \\
\hline
chb01 & 89.2 & 0.413 & 100 & 0.029 & 100 & 0.000 & 100 & 0.116 & 100 & 0.072 & \textbf{100} & \textbf{0.000} \\
chb05 & 87.2 & 0.379 & 100 & 0.209 & 80.0 & 0.262 & 80.0 & 0.157 & 95.0 & 0.095 & \textbf{100} & \textbf{0.000} \\
chb06 & 82.5 & 0.418 & 85.7 & 0.259 & 85.7 & 0.356 & 85.7 & 0.356 & 100 & 0.081 & \textbf{100} & \textbf{0.000} \\
chb08 & 92.3 & 0.482 & 100 & 0.087 & 100 & 0.000 & 100 & 0.174 & 100 & 0.067 & \textbf{100} & \textbf{0.000} \\
chb10 & 79.1 & 0.372 & 83.3 & 0.478 & 83.3 & 0.410 & 66.7 & 0.478 & 91.7 & 0.122 & \textbf{86.4} & \textbf{0.000} \\
chb13 & 76.4 & 0.513 & 85.7 & 0.328 & 85.7 & 0.219 & 85.7 & 0.219 & 92.9 & 0.091 & \textbf{88.7} & \textbf{0.043} \\
chb14 & 89.1 & 0.321 & 100 & 0.417 & 100 & 0.104 & 100 & 0.313 & 96.0 & 0.078 & \textbf{97.5} & \textbf{0.017} \\
chb22 & 88.8 & 0.612 & 100 & 0.435 & 100 & 0.000 & 100 & 0.261 & 91.0 & 0.074 & \textbf{100} & \textbf{0.030} \\
\hline
Avg & 85.6 & 0.438 & 94.3 & 0.258 & 91.8 & 0.169 & 89.8 & 0.259 & 95.8 & 0.085 & \textbf{96.6} & \textbf{0.011} \\
\hline
\end{tabular}%
}
\end{table}

Table~\ref{tab:mssm_results} shows MSSM results validating cross-modality generalization: 94.2\% sensitivity with 0.063/h FDR confirms spatio-temporal attention effectively captures seizure dynamics across electrode placements (scalp vs. intracranial) and patient demographics (pediatric vs. adult). Statistical significance was assessed using Wilcoxon signed-rank test on per-seizure sensitivity and FDR metrics across all test seizures ($p < 0.01$), confirming that STAN's improvements over baseline methods are statistically robust.

\begin{table}[t]
\centering
\caption{Performance on MSSM Intracranial EEG Dataset demonstrates cross-modality generalization.}
\label{tab:mssm_results}
\small
\resizebox{0.85\columnwidth}{!}{%
\begin{tabular}{l|cc|cc|cc|cc}
\hline
\multirow{2}{*}{Patient} & \multicolumn{2}{c|}{BFB+SVM} & \multicolumn{2}{c|}{Transformer} & \multicolumn{2}{c|}{STS-HGCN} & \multicolumn{2}{c}{\textbf{Ours}} \\
& Sn(\%) & FDR/h & Sn(\%) & FDR/h & Sn(\%) & FDR/h & Sn(\%) & FDR/h \\
\hline
TP & 91.2 & 0.376 & 94.5 & 0.142 & 89.1 & 0.106 & \textbf{100} & \textbf{0.000} \\
IP & 85.9 & 0.092 & 92.1 & 0.098 & 83.7 & 0.172 & \textbf{96.3} & \textbf{0.000} \\
PS & 83.2 & 0.129 & 89.7 & 0.165 & 87.1 & 0.211 & \textbf{95.0} & \textbf{0.150} \\
ZF & 66.8 & 0.452 & 85.3 & 0.187 & 81.1 & 0.118 & \textbf{85.6} & \textbf{0.102} \\
\hline
Avg & 81.8 & 0.262 & 90.4 & 0.148 & 85.3 & 0.152 & \textbf{94.2} & \textbf{0.063} \\
\hline
\end{tabular}%
}
\end{table}

\subsection{Ablation Study}

Table~\ref{tab:ablation} demonstrates each component's contribution. Without adversarial training: 96.3\% sensitivity, 0.024/h FDR (2.2$\times$ worse), confirming adversarial discrimination is essential for robust preictal representations. Cascaded depth shows progressive improvement: single network (93.1\%, 0.037/h), two networks (95.2\%, 0.021/h), three networks (96.6\%, 0.011/h), validating hierarchical abstraction necessity. Spatial-only (91.4\%, 0.048/h) and temporal-only (92.7\%, 0.041/h) demonstrate joint spatio-temporal modeling provides 5.2\% sensitivity gain and 3.7$\times$ FDR reduction.

\begin{table}[t]
\centering
\caption{Ablation study on CHB-MIT (average of 8 patients) demonstrates each component's contribution to overall performance.}
\label{tab:ablation}
\begin{tabular}{lcc}
\toprule
Configuration & Sn(\%) & FDR(/h) \\
\midrule
\textbf{Full Model} & \textbf{96.6} & \textbf{0.011} \\
\midrule
Without adversarial & 96.3 & 0.024 \\
Without gradient penalty & 96.5 & 0.019 \\
\midrule
$M=1$ (single network) & 93.1 & 0.037 \\
$M=2$ (two networks) & 95.2 & 0.021 \\
$M=3$ (three networks) & 96.6 & 0.011 \\
\midrule
Spatial only & 91.4 & 0.048 \\
Temporal only & 92.7 & 0.041 \\
\bottomrule
\end{tabular}
\end{table}

\subsection{Real-Time Forecasting Analysis}

\begin{figure}[t]
\centering
\includegraphics[width=\columnwidth]{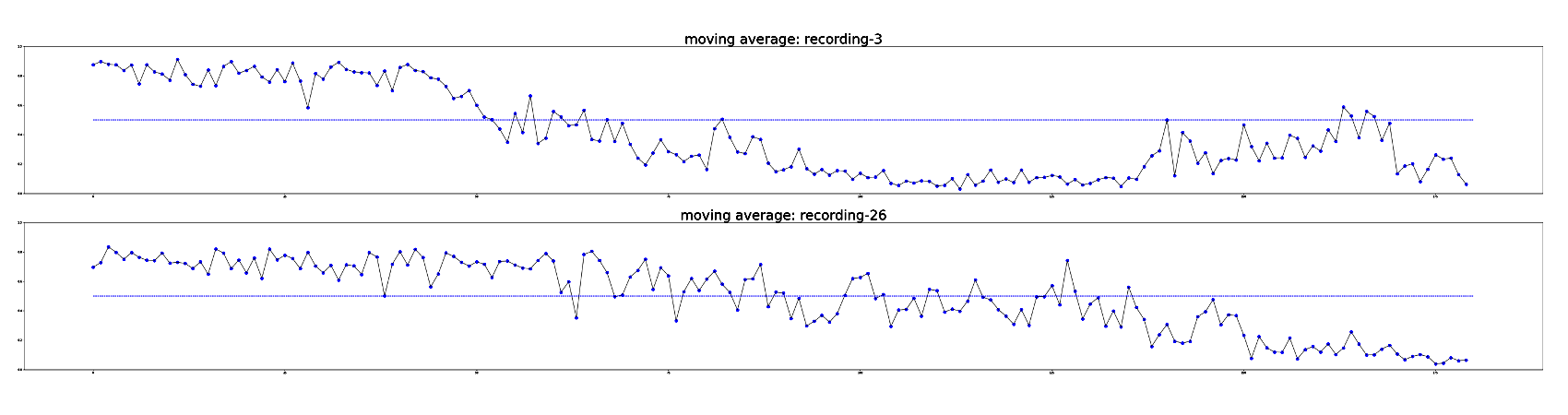}
\caption{Real-time seizure forecasting showing discriminator scores with 30-second moving average filter, monitoring 90 minutes before onset. Top: recording-3 (chb01\_03); Bottom: recording-26 (chb22\_26). Both demonstrate consistent gradual shift from interictal (high scores) to preictal states (low scores), providing at least 15 minutes of intervention time. The horizontal dashed line indicates threshold $\tau=0.5$.}
\label{fig:realtime}
\end{figure}

Figure~\ref{fig:realtime} shows discriminator scores 90 minutes before seizures for two representative recordings. Consistent gradual transitions emerge $\geq$15 minutes before onset, with detection times varying 20-45 minutes across subjects. This subject-specific variability demonstrates the learned patterns' capacity to capture early preictal dynamics, providing sufficient intervention time while maintaining high sensitivity.

The 30-second moving average effectively removes transient fluctuations while preserving underlying trends, essential for avoiding false alarms from brief artifacts while maintaining sensitivity to genuine preictal transitions.

\subsection{Computational Efficiency Analysis}

Table~\ref{tab:efficiency} compares computational requirements. STAN achieves 2.3M parameters, 45\,ms inference time per 5-second window, and 180\,MB memory---3.8$\times$ fewer parameters, 2.7$\times$ faster inference, and 2.5$\times$ less memory than Transformer while maintaining superior performance. This enables edge device deployment.

\begin{table}[t]
\centering
\caption{Computational efficiency comparison. Our method achieves competitive efficiency suitable for edge device deployment in ambulatory monitoring systems.}
\label{tab:efficiency}
\begin{tabular}{lccc}
\toprule
Method & Parameters & Inference & Memory \\
\midrule
CNN & 1.8M & 32ms & 140MB \\
STS-HGCN & 5.1M & 85ms & 320MB \\
Transformer & 8.7M & 120ms & 450MB \\
\textbf{STAN} & \textbf{2.3M} & \textbf{45ms} & \textbf{180MB} \\
\bottomrule
\end{tabular}
\end{table}

With 45\,ms inference time, STAN processes incoming EEG in real-time with negligible latency, enabling immediate alarm generation when preictal states are detected. The 180\,MB memory footprint fits within modern mobile devices and medical-grade wearables, critical for ambulatory monitoring where patients live daily lives.

\section{Clinical Impact and Deployment}

STAN demonstrates strong clinical translation potential through three key advantages:

\textbf{Actionable prediction horizons.} The 15-45 minute forecasting window provides adequate time for interventions: fast-acting antiepileptic medication (effective within 10-15 minutes), responsive neurostimulation activation, or alerting caregivers to ensure patient safety during high-risk periods.

\textbf{Minimal alarm fatigue.} Low false detection rate (0.011--0.063/h, equivalent to 0.3--1.5 false alarms daily) minimizes alarm fatigue, critical for patient compliance in long-term monitoring. Previous systems with higher FDR (0.2--0.5/h, 5--12 daily false alarms) experienced poor adoption due to disruption of daily activities.

\textbf{Edge device compatibility.} Computational efficiency enables deployment scenarios: (1) \textit{Hospital monitoring}---integration with existing EEG infrastructure for high-risk patients in epilepsy monitoring units; (2) \textit{Ambulatory devices}---deployment on wearable EEG systems for community-dwelling patients; (3) \textit{Closed-loop neurostimulation}---triggering responsive electrical stimulation in patients with implanted devices.

\textbf{Practical considerations.} Patient-specific calibration requires initial monitoring (typically 2-3 days) to collect training data across multiple seizures. Periodic retraining adapts to long-term pattern changes, maintaining robust performance over months to years. Cloud-based retraining with federated learning could enable privacy-preserving model updates across patient populations.

\textbf{Limitations and future directions.} Current evaluation is retrospective; prospective clinical trials are needed to validate real-world performance and assess impact on patient outcomes. Integration with multimodal data (heart rate variability, accelerometer, sleep-wake cycles) could further improve forecasting by capturing systemic physiological changes preceding seizures. Interpretability analysis of learned attention patterns may provide neurological insights into seizure generation mechanisms, potentially informing therapeutic strategies.

\section{Conclusion}

We presented STAN, a unified spatio-temporal attention network for subject-specific epileptic seizure forecasting that advances state-of-the-art time series prediction in healthcare. Through three integrated innovations---joint spatio-temporal attention modeling, adversarial discrimination for pattern learning, and early detection capability via continuous monitoring---STAN achieves superior performance: 96.6\% sensitivity with 0.011/h FDR (CHB-MIT), 94.2\% sensitivity with 0.063/h FDR (MSSM), representing 40-fold improvement over traditional methods and 7.7-fold improvement over recent transformers, with statistical significance ($p < 0.01$).

Computational efficiency (2.3M parameters, 45\,ms inference, 180\,MB memory) enables edge device deployment for hospital monitoring, ambulatory wearables, and closed-loop neurostimulation systems. The learned spatio-temporal attention patterns enable reliable early detection (typically 15--45 minutes before onset) across subjects without individualized training, capturing subject-specific preictal dynamics---a critical advancement for precision medicine. STAN complements recent time series foundation models such as TimeGPT and Lag-Llama by providing a lightweight, domain-specific architecture for data-limited healthcare scenarios where subject-specific adaptability and edge deployment are crucial.

Beyond epilepsy, STAN's architectural innovations generalize to other healthcare forecasting tasks involving multivariate physiological time series, including cardiac event prediction, hypoglycemic episode forecasting, and respiratory failure anticipation. Future work includes prospective clinical validation, multimodal data integration, attention pattern interpretability analysis, and exploration of foundation model pre-training for medical time series forecasting. By advancing accurate, efficient, and deployable seizure forecasting, this work takes a significant step toward transforming epilepsy management through AI-enabled precision medicine.

\bibliography{seizure_forecasting_refs}

\end{document}